\documentclass{article}

\usepackage{microtype}
\usepackage{graphicx}
\usepackage{booktabs}
\usepackage{hyperref}
\usepackage{subcaption}

\usepackage[accepted]{icml2025}
\usepackage{amsmath}
\usepackage{amssymb}
\usepackage{mathtools}
\usepackage{amsthm}
\usepackage[capitalize,noabbrev]{cleveref}

\theoremstyle{plain}

\theoremstyle{definition}

\theoremstyle{remark}

\usepackage[textsize=tiny]{todonotes}

\icmltitlerunning{When Does Routing Become Interpretable?}

\begin{document}







\twocolumn[
\icmltitle{When Does Routing Become Interpretable?\\
          Causal Probes on Block Attention Residuals}

\vskip 0.15in
\begin{center}
{\bf Aydin Javadov}\\
ETH Zurich\\
\end{center}

\icmlkeywords{Machine Learning, Mechanistic Interpretability, Residual Connections, Attention}

\vskip 0.3in
]

\begin{abstract}
Block Attention Residuals (Block AttnRes) by \cite{Chen2026AttentionR} replace fixed additive residuals \cite{DBLP:journals/corr/HeZRS15} with a learned softmax over earlier depth-source representations, surfacing cross-layer routing as an inspectable tensor in the forward pass. This is a tempting interpretability target: information flow normally inferred indirectly is now directly observable. We ask whether such exposure suffices for mechanistic interpretation. We probe two same-scale ($0.6$B) Block AttnRes checkpoints under identical routing-ablation interventions: a vanilla Qwen3 inference-wrapped through a deterministic recency-bias schedule that the codebase admits as a routing-equivalent loading path, and a Block AttnRes Qwen3 trained from scratch with routing as part of optimisation. The wrapped baseline's routing weights are content-independent and reproduce the schedule's analytic prediction. The trained AttnRes checkpoint instead exhibits three localised routing motifs: an embedding-source pathway through early-layer MLP, a current-state pathway through early-layer attention and MLP, and an older-history pathway through late-layer attention. Beyond this stratification, we find a sharp dissociation between average routing mass and causal importance: in both sublayers, the largest mass slice is not the largest causal contribution, and one source family carries appreciable mass with no detectable causal role under intervention. Architectural exposure of routing is therefore necessary but not sufficient for mechanistic interpretation: structured depth routing emerges only when routing has been part of training, and even then, descriptive routing summaries should be treated as candidate hypotheses to be tested by causal interventions, not as evidence of mechanism in their own right. 
\newpage Code is available at \url{https://anonymous.4open.science/r/attn_res_interp-E70C/}.

\end{abstract}

\section{Introduction}

Cross-layer information flow is the central object of mechanistic interpretability work on Transformers \citep{rai2024practical}, but it is rarely an explicit object in the forward pass. Standard PreNorm Transformers \citep{DBLP:journals/corr/abs-2002-04745} accumulate sublayer outputs additively with fixed unit weights, and any structure in how depth states combine has to be recovered indirectly through patching, probing, or path analysis \citep{meng2022locating}. \citet{Chen2026AttentionR} introduce Attention Residuals (AttnRes) and their scalable variant Block AttnRes  change this: each sublayer recombines earlier depth-source representations through a learned softmax over depth, so the routing weights $\alpha_{i\to\ell}$ are a tensor that can be read off the forward pass.

This raises a natural interpretability question: with depth routing now made explicit, can cross-layer information flow be \emph{read} rather than \emph{inferred}? We sharpen it: can one take a pretrained vanilla Transformer, expose its forward pass through the AttnRes class, and read mechanism off the resulting routing weights? We build on a publicly released codebase \footnote{https://github.com/wdlctc/open-attention-residuals} that admits this construction. With per-sublayer query projection set to zero and a positive bias on the current-source logit, the depth-softmax reduces to a deterministic schedule that puts most weight on the current source and the rest uniformly across earlier ones. Under this construction, a vanilla checkpoint can be loaded into the AttnRes class and produce routing weights, even though those weights carry no learned content.

\begin{figure}[t]
\centering
\includegraphics[width=0.95\linewidth]{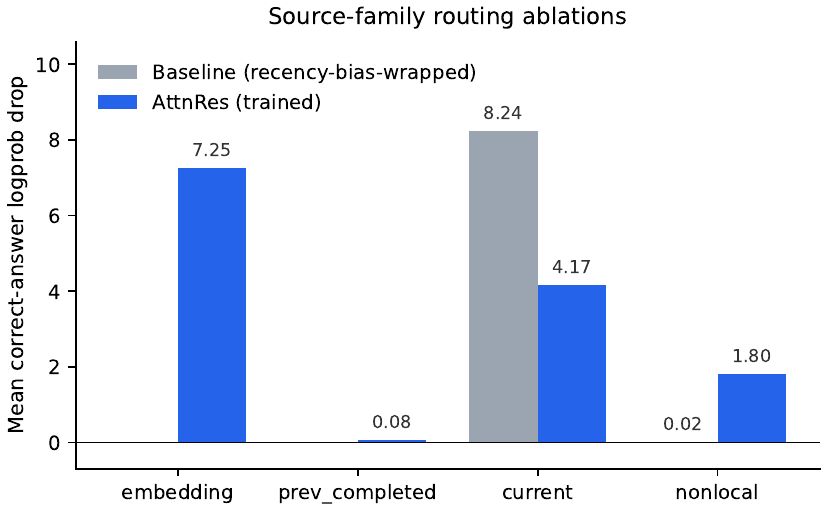}
\caption{Source-family routing ablations: baseline (recency-bias-wrapped vanilla Qwen3) vs.\ Block AttnRes trained from scratch, on the same probe and the same data. Three families produce strong AttnRes-side effects with negligible baseline counterparts; a fourth carries routing mass but no causal weight.}
\label{fig:headline}
\end{figure}

We argue that interpreting routing weights produced by this kind of post-hoc wrapping is uninformative by construction, and we make the argument concrete by running identical routing-ablation probes on two $0.6$B checkpoints from the repository: a \emph{baseline} that is a vanilla Qwen3 trained with standard additive residuals and inference-wrapped through the recency-bias schedule, and an \emph{AttnRes} model that is a Block AttnRes Qwen3 trained from scratch with the routing as part of optimisation. Both expose the same routing-weight tensors to the same probe (Figure~\ref{fig:headline}).

Three findings follow. First, the baseline routing pattern reproduces the recency-bias schedule's analytic prediction: mass on the current source averages $0.840$, attention and MLP rows of the routing-mass table are identical, and only ablation of the current-state MLP pathway is causally consequential. Second, the AttnRes checkpoint exhibits three localised motifs in the (source-family, sublayer, depth) cube --- embedding through early-layer MLP, current through early-layer attention and MLP, and older nonlocal history through late-layer attention --- with a fourth source family carrying appreciable routing mass but no detectable causal role. Third, average routing mass and causal importance dissociate sharply on the AttnRes side: in both sublayers the largest mass slice is not the largest causal contribution.

\paragraph{Contributions.}
\begin{itemize}\setlength{\itemsep}{0.2em}
    \item A routing-ablation framework for Block AttnRes that masks mutually exclusive source families and renormalises the remainder, jointly factored by sublayer and depth region.
    \item A controlled demonstration that explicit depth routing alone is insufficient for mechanistic interpretation: under matched probes, a recency-bias-wrapped vanilla checkpoint exposes routing weights that reproduce the wrapping schedule's analytic prediction, while a Block AttnRes checkpoint trained from scratch exhibits three localised causal motifs.
    \item Evidence that average routing mass and causal importance dissociate sharply on the AttnRes side, including one source family that carries appreciable mass with no detectable causal role.
\end{itemize}

\section{Background: Attention Residuals}

In standard PreNorm Transformers, the residual update for sublayer $f_\ell$ at depth $\ell$ is
\begin{equation}
h_{\ell+1} = h_\ell + f_\ell(h_\ell),
\end{equation}
unrolling as a fixed-coefficient sum over earlier states with each contribution at unit weight. The original AttnRes paper \citep{Chen2026AttentionR} argues that this produces \emph{PreNorm dilution} \citep{Li2026SiameseNormBT} and replaces the sum with a learned softmax over earlier depth states,
\begin{equation}
h_{\ell+1} = \sum_{i\le\ell} \alpha_{i\to\ell}\, h_i, \qquad \sum_{i\le\ell}\alpha_{i\to\ell} = 1,
\end{equation}
where $\alpha_{i\to\ell}$ is content-dependent and computed by a per-sublayer projection. To control memory, the released code uses \emph{Block AttnRes}, in which routing operates over a compressed source axis: completed-block summaries followed by the current partial block. For an $L$-layer model partitioned into $B$ blocks, the source axis at a sublayer call has up to $B+1$ entries; in the $0.6$B AttnRes checkpoint we study ($L=28$, $B=8$), sublayer calls occur with source-axis lengths $S\in\{2,\ldots,8\}$.

\paragraph{The recency-bias schedule.}
The released implementation adds a learnable scalar bias $\beta$ to the partial-block logit before the depth-softmax. With per-sublayer query projection $\mathbf{W}=\mathbf{0}$ and $\beta>0$, the depth-softmax reduces to a deterministic schedule
\begin{equation}
\alpha = \Big[\tfrac{1}{S{-}1+e^\beta},\dots,\tfrac{1}{S{-}1+e^\beta},\tfrac{e^\beta}{S{-}1+e^\beta}\Big],
\label{eq:recency}
\end{equation}
a fixed convex combination of the current state and a uniform mixture of earlier sources. This schedule does not depend on the input and is the routing pattern the codebase uses for converting pretrained checkpoints into the AttnRes class. We use it as a control: a routing pattern that exists architecturally but carries no learned content.

While the original AttnRes paper's contribution is architectural \citep{Chen2026AttentionR}, our concern is more on: given that Block AttnRes makes depth routing an explicit learned object, what do its routing weights actually do, and is the architectural exposure of routing sufficient for interpretation?

\section{Method}

\subsection{Task}

We construct a simple design by synthetic key--value retrieval prompts. Each example presents a short list of distinct keys mapped to distinct values, followed by a query key, and asks the model to score four candidate continuations. Three distractors are drawn preferentially from values present in the same prompt; the correct value is the one mapped to the query key. Continuations are scored by their conditional log-probability under the model. Across $200$ examples we cycle through three prompt templates in fixed proportion to reduce surface-form sensitivity. Full dataset details are in Appendix~\ref{app:dataset}.

\subsection{Source families}

The Block AttnRes source axis at a sublayer call has length $S$, ordering completed-block summaries first and the current partial block last. We group these indices into four mutually exclusive families: \textbf{embedding}, the first source; \textbf{current}, the partial block; \textbf{prev\_completed}, the most recent completed block summary, distinct from the embedding source; and \textbf{nonlocal}, older completed blocks excluding the most recent. The families are disjoint at every $S$; for the smallest source-axis lengths some families are empty by definition (e.g.\ \texttt{nonlocal} requires $S \ge 4$). Full definitions are in Appendix~\ref{app:ablations}.

\subsection{Routing ablations}

Our central intervention is a \emph{mask-and-renormalise} ablation. Given a target family $\mathcal{F}$ and a softmax-normalised routing vector $\alpha\in\mathbb{R}^S$, we form a binary mask $m\in\{0,1\}^S$ with $m_s=0$ iff $s\in\mathcal{F}$, and replace $\alpha$ with
\begin{equation}
\tilde\alpha_s = \frac{m_s\,\alpha_s}{\sum_{s'} m_{s'}\,\alpha_{s'}}.
\end{equation}
The total mass entering the sublayer is preserved; what changes is the model's reliance on the targeted pathway. \texttt{keep\_only}-style interventions invert the mask. A single ablation can be restricted along three axes simultaneously: source family, sublayer $\in\{\textsc{attn},\textsc{mlp}\}$, and layer subset (early, middle, late thirds of the residual stack).

\subsection{Two checkpoints, one probe}

We run identical probes on two $0.6$B checkpoints from the publicly released codebase mentioned before.

\paragraph{AttnRes condition.} A Block AttnRes Qwen3 ($d{=}1024$, $L{=}28$, $B{=}8$) trained from scratch with routing parameters as part of optimisation.

\paragraph{Baseline condition.} A vanilla Qwen3 of identical scale, trained with standard additive residuals and loaded for evaluation through the same AttnRes model class. In every layer of the loaded checkpoint, the per-sublayer query projection has zero norm and the recency bias is at the codebase default $\beta=3.0$, so by Equation~\ref{eq:recency} the routing weights are a deterministic function of the source-axis length and a fixed scalar; no input-dependent routing decisions are made. The wrapping is not output-preserving (Appendix~\ref{app:checkpoints}); the property we need for the control is that the routing is content-independent.

This comparison controls for the probe, the trace mechanism, the model class, scale, tokenizer, and dataset. Differences between the two conditions should be interpreted as differences between a content-independent routing surface introduced at evaluation time and routing parameters optimised during training.

\subsection{Metrics}

We report mean accuracy on the 4-choice retrieval task; mean drop in correct-answer total log-probability relative to the unablated checkpoint (\textit{logprob $\downarrow$}); fraction of examples whose argmax prediction flips relative to that unablated state; and mean rank of the correct answer among the four candidates. All numbers are computed on the same $200$-example dataset for both checkpoints.

\section{Results}

\subsection{Baseline routing matches the recency-bias schedule}
\label{sec:baseline-degenerate}

Source-family ablations on the recency-bias-wrapped baseline (Table~\ref{tab:source-base}) match the schedule's analytic prediction. Ablating \texttt{embedding}, \texttt{prev\_completed}, or \texttt{nonlocal} sources leaves correct-answer log-probability essentially unchanged ($|\Delta| < 0.025$); these families collectively receive only $\approx 0.16$ of the routing mass through the uniform-residual term of Equation~\ref{eq:recency}, so reweighting among them does little. Ablating \texttt{current} drops accuracy from $0.560$ to $0.305$ and produces a logprob drop of $8.24$, since that source carries the bulk of the mass.

\begin{table}[t]
\centering
\footnotesize
\setlength{\tabcolsep}{4pt}
\caption{Source-family routing ablations on the baseline. Apart from \texttt{current}, family ablations are no-ops, as predicted by the recency-bias schedule (Eq.~\ref{eq:recency}).}
\label{tab:source-base}
\begin{tabular}{lrrrr}
\toprule
Ablation & Acc. & Logprob $\downarrow$ & Flips & Rank \\
\midrule
none                   & 0.560 &  0.000 & 0.000 & 1.795 \\
drop\_embedding        & 0.555 &  0.001 & 0.005 & 1.805 \\
drop\_prev\_completed  & 0.565 & -0.001 & 0.005 & 1.790 \\
drop\_nonlocal         & 0.555 &  0.020 & 0.025 & 1.805 \\
drop\_current          & 0.305 &  8.240 & 0.765 & 2.400 \\
keep\_only\_embedding  & 0.225 & 27.675 & 0.860 & 2.515 \\
keep\_only\_current    & 0.555 &  0.015 & 0.030 & 1.805 \\
keep\_only\_completed  & 0.235 & 10.914 & 0.800 & 2.525 \\
\bottomrule
\end{tabular}
\end{table}

The descriptive routing-mass summary (Table~\ref{tab:mass}) confirms the mechanism: mass on \texttt{current} is $0.840$ in both sublayers, matching the analytic prediction $\overline{e^\beta/(S{-}1+e^\beta)} = 0.840$ over the model's depth schedule (Appendix~\ref{app:checkpoints}); attention and MLP rows are identical to four decimal places, which is consistent only with both per-sublayer projections being inert. The baseline routing pattern is the recency-bias schedule, and depth-aware probes recover nothing the schedule does not predict on paper.

\subsection{AttnRes shows differentiated source-family causality}
\label{sec:attnres-source}

The same probes on the AttnRes checkpoint produce an entirely different picture (Table~\ref{tab:source-attnres}). \texttt{embedding} is the largest contributor (logprob drop $7.25$), exceeding \texttt{current} ($4.17$) by nearly a factor of two. Older \texttt{nonlocal} history contributes a smaller but clearly measurable effect ($1.80$). \texttt{prev\_completed} ablation produces effectively no change ($0.08$), despite this family carrying $\approx 0.09$ of the routing mass on average (Section~\ref{sec:mass-importance}). Base-rate accuracy is comparable to the baseline ($0.540$ vs.\ $0.560$): the contrast in which sources matter reflects a different mechanism, not a different competence.

\begin{table}[t]
\centering
\footnotesize
\setlength{\tabcolsep}{4pt}
\caption{Source-family routing ablations on the AttnRes checkpoint. \texttt{embedding} produces the largest drop, followed by \texttt{current} and \texttt{nonlocal}. \texttt{prev\_completed} ablation has no detectable causal effect.}
\label{tab:source-attnres}
\begin{tabular}{lrrrr}
\toprule
Ablation & Acc. & Logprob $\downarrow$ & Flips & Rank \\
\midrule
none                   & 0.540 &  0.000 & 0.000 & 1.760 \\
drop\_embedding        & 0.235 &  7.250 & 0.810 & 2.570 \\
drop\_current          & 0.325 &  4.175 & 0.665 & 2.215 \\
drop\_nonlocal         & 0.470 &  1.800 & 0.280 & 1.995 \\
drop\_prev\_completed  & 0.545 &  0.076 & 0.090 & 1.785 \\
keep\_only\_embedding  & 0.215 & 24.978 & 0.840 & 2.620 \\
keep\_only\_current    & 0.240 &  7.409 & 0.820 & 2.520 \\
keep\_only\_completed  & 0.175 &  7.256 & 0.805 & 2.650 \\
\bottomrule
\end{tabular}
\end{table}

The contrast against the baseline is sharp on the same metric: \texttt{drop\_embedding} goes from $0.001$ to $7.25$, \texttt{drop\_nonlocal} from $0.020$ to $1.80$, and \texttt{drop\_current} from $8.24$ down to $4.17$ as the current-specific dependence is partially redistributed to other families.

\subsection{Sublayer asymmetry only emerges under AttnRes}

\begin{figure}[t]
\centering
\includegraphics[width=0.95\linewidth]{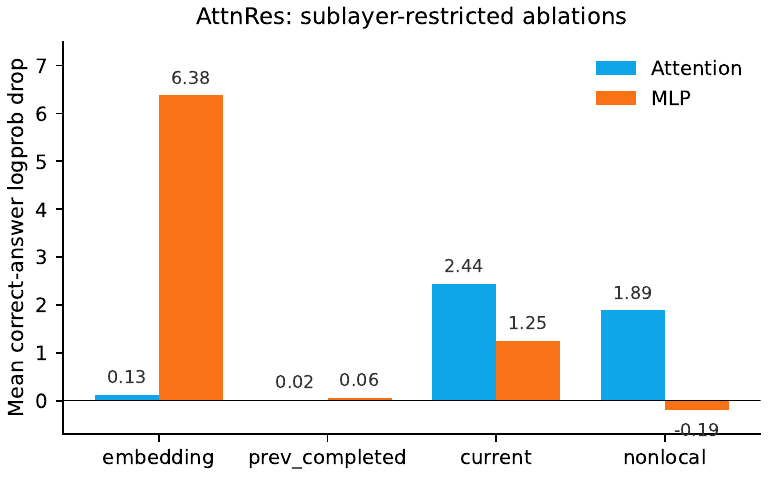}
\caption{Sublayer-restricted routing ablations on the AttnRes checkpoint. \texttt{embedding} matters only through MLP; \texttt{nonlocal} only through attention; \texttt{current} through both, with attention larger; \texttt{prev\_completed} through neither.}
\label{fig:sublayer}
\end{figure}

Splitting routing ablations by sublayer locates where in the per-layer computation each source family matters (Table~\ref{tab:sublayer}, Figure~\ref{fig:sublayer}). The baseline has only one non-trivial row; its split between attention and MLP for \texttt{current} reflects which sublayer the recency-bias-dominated routing is most fragile to perturbation in, not learned structure. The AttnRes checkpoint differentiates each row: \texttt{embedding} matters almost entirely through MLP (drop $6.38$ vs.\ $0.13$); \texttt{current} reverses the baseline order, with attention ($2.44$) exceeding MLP ($1.25$); \texttt{nonlocal} matters only through attention; and \texttt{prev\_completed} is below noise in both. Sublayer asymmetry is a property of the trained model, not of the probe.

\begin{table}[t]
\centering
\footnotesize
\setlength{\tabcolsep}{4pt}
\caption{Sublayer-restricted routing ablations: mean correct-answer logprob drop. The baseline shows a single non-trivial cell (current-MLP); the AttnRes checkpoint stratifies across sublayers in a family-specific way.}
\label{tab:sublayer}
\begin{tabular}{lrrrr}
\toprule
& \multicolumn{2}{c}{Baseline} & \multicolumn{2}{c}{AttnRes} \\
\cmidrule(lr){2-3}\cmidrule(lr){4-5}
Source family   & Attn & MLP   & Attn & MLP   \\
\midrule
embedding       &  0.001 & -0.001 & 0.130 & 6.382 \\
prev\_completed & -0.006 &  0.003 & 0.016 & 0.064 \\
current         &  2.643 &  8.824 & 2.444 & 1.254 \\
nonlocal        &  0.043 & -0.018 & 1.885 & -0.193 \\
\bottomrule
\end{tabular}
\end{table}

\subsection{Routing mass and causal importance dissociate}
\label{sec:mass-importance}

Average routing weights, taken descriptively, do not predict the causal results. Table~\ref{tab:mass} reports average source-family mass at the final answer token on correctly-solved examples.

\begin{table}[t]
\centering
\footnotesize
\setlength{\tabcolsep}{4pt}
\caption{Average final-token routing mass on correctly-solved examples. Baseline rows are identical and dominated by \texttt{current}. AttnRes attention is broadest with the largest slice on \texttt{nonlocal}; AttnRes MLP is current-heavy. Each row sums to $\approx 1.0$.}
\label{tab:mass}
\begin{tabular}{llrrrr}
\toprule
Model    & Sublayer & Emb. & Prev. & Nonloc. & Curr. \\
\midrule
Baseline & attn & 0.042 & 0.035 & 0.083 & 0.840 \\
Baseline & mlp  & 0.042 & 0.035 & 0.083 & 0.840 \\
AttnRes  & attn & 0.195 & 0.099 & 0.368 & 0.338 \\
AttnRes  & mlp  & 0.228 & 0.085 & 0.083 & 0.603 \\
\bottomrule
\end{tabular}
\end{table}

On the AttnRes side, the largest mass slice in each sublayer (\texttt{nonlocal} in attention at $0.368$; \texttt{current} in MLP at $0.603$) is not the largest causal contribution: the corresponding ablations cost $1.89$ and $1.25$ logprob, while \texttt{drop\_current\_attn} costs $2.44$ and \texttt{drop\_embedding\_mlp} costs $6.38$. Embedding in MLP carries less than half of current's mass yet five times its causal effect. And \texttt{prev\_completed}, with mass $0.085$--$0.099$ per sublayer comparable to \texttt{nonlocal} in MLP, has no detectable causal role at all (drops $0.016$ and $0.064$). Even when the routing pattern is structured, average mass is a misleading indicator of causal importance.

\subsection{The AttnRes structure is depth-localised}

Binning layers into early, middle, and late thirds and repeating the sublayer ablations gives Table~\ref{tab:layerbin} and Figure~\ref{fig:cube}; the full table is in Appendix~\ref{app:layerbin}.

\begin{table}[t]
\centering
\footnotesize
\setlength{\tabcolsep}{4pt}
\caption{Layer-binned sublayer ablations on the AttnRes checkpoint, mean correct-answer logprob drop. Three localised motifs dominate the causal signal: \texttt{embedding} early-MLP, \texttt{current} early-attention and early-MLP, \texttt{nonlocal} late-attention.}
\label{tab:layerbin}
\begin{tabular}{llrrr}
\toprule
Source family & Sublayer & Early & Middle & Late \\
\midrule
embedding       & attn & 0.140 & -0.003 & -0.000 \\
embedding       & mlp  & 6.381 & -0.000 &  0.002 \\
prev\_completed & attn & 0.045 & -0.040 &  0.008 \\
prev\_completed & mlp  & 0.011 &  0.008 &  0.050 \\
current         & attn & 2.321 & -0.120 &  0.085 \\
current         & mlp  & 1.122 &  0.494 &  0.116 \\
nonlocal        & attn & 0.076 & -0.079 &  1.729 \\
nonlocal        & mlp  & 0.005 &  0.018 & -0.202 \\
\bottomrule
\end{tabular}
\end{table}

\begin{figure}[t]
\centering
\includegraphics[width=0.95\linewidth]{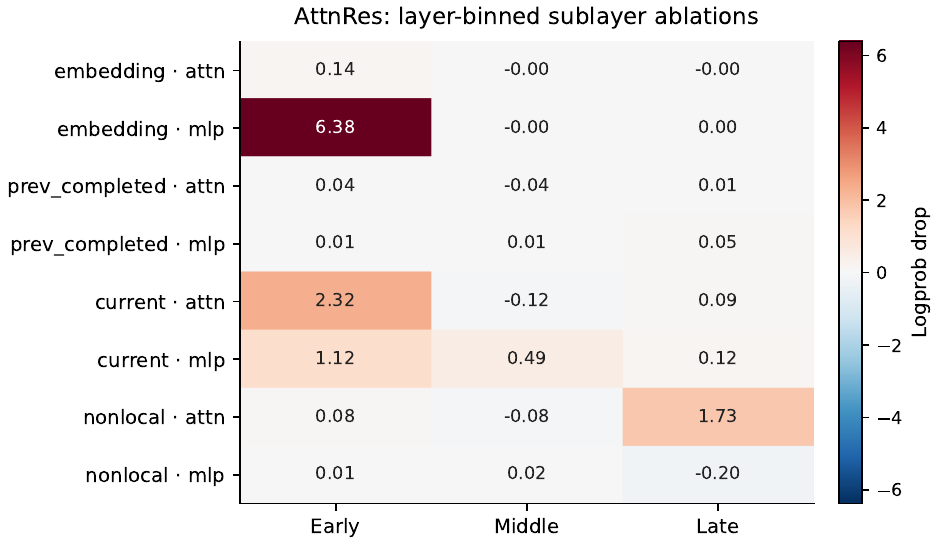}
\caption{Layer-binned sublayer ablations on the AttnRes checkpoint as a heatmap. A small number of localised cells carry the primary causal signal.}
\label{fig:cube}
\end{figure}

The entire \texttt{embedding} effect sits in early-layer MLP routing ($6.38$, near zero elsewhere). \texttt{current} concentrates early, with attention ($2.32$) exceeding MLP ($1.12$) --- the only family for which attention exceeds MLP in the early bin. \texttt{nonlocal} appears only in late attention ($1.73$). \texttt{prev\_completed} is below noise everywhere. The picture is of a stage-specific routing protocol: early MLP as the injection point for retrieved depth content from the embedding source, early attention as the read-out point for the current working state with a smaller MLP-side counterpart, and late attention as occasional long-reach access into older completed history.

\section{Discussion}

The \citep{Chen2026AttentionR} showed that learned content-dependent routing across depth is useful for optimisation and downstream performance. We complement that by characterising \emph{when} the routing it exposes becomes mechanistically interpretable, and by what method.

\paragraph{Architectural exposure is necessary but not sufficient.} The recency-bias-wrapped baseline is, formally, a Block AttnRes model: same forward function, same routing tensors, same model class. Its routing parameters are nevertheless inert, so by Equation~\ref{eq:recency} its routing pattern is a deterministic function of source-axis length and a fixed scalar. The pattern we observe matches that prediction precisely. Block AttnRes does not, by construction, render cross-layer information flow legible; it provides a substrate on which a model that has been trained to use depth diversity can express that usage in a directly intervenable form.

\paragraph{Three localised motifs.} Of the $24$ (family, sublayer, depth) cells in the AttnRes ablation cube, three localised motifs dominate the causal signal --- early-layer MLP for \texttt{embedding}, early-layer attention and MLP for \texttt{current}, late-layer attention for \texttt{nonlocal}. This is consistent with early MLP as an injection point for the embedding-origin source, early attention as a read-out point for the current working state, and late attention as occasional long-reach access into older completed history. We make no claim that this is the only solution Block AttnRes admits; it is the solution this trained checkpoint converged to, recoverable by our probes precisely because the routing parameters are non-degenerate.

\paragraph{Mass and causal importance must be separated.} Three distinct dissociations show up in the AttnRes checkpoint: large mass with small causal effect (\texttt{nonlocal} in attention; \texttt{current} in MLP), small mass with large effect (\texttt{embedding} in MLP), and appreciable mass with no detectable effect at all (\texttt{prev\_completed}). Routing-weight visualisations alone are not sufficient evidence of mechanism.

\paragraph{Implications.} Two for routing-exposing architectures more generally. First, exposure of routing weights does not transfer across the training boundary: a vanilla model wrapped to make routing visible will surface weights that carry no information beyond what the wrapping schedule dictates. Second, even when routing has been trained, average weights are best treated as candidate pathways, with causal interventions deciding which carry decision-critical signal.

\section{Limitations}

\begin{itemize}\setlength{\itemsep}{0.2em}
    \item \textbf{Task simplicity.} Synthetic key--value retrieval rather than naturalistic long-context reasoning. The structure may not transfer.
    \item \textbf{Two regimes only.} A single baseline and a single AttnRes checkpoint, single scale, single recipe. We have not run intermediate conditions (AttnRes architecture fine-tuned from a vanilla checkpoint, or trained with routing parameters frozen), so we cannot strictly attribute the structured pattern to training-from-scratch as opposed to AttnRes training in general.
    \item \textbf{Block-level routing.} Our interventions operate on block summaries; full per-layer AttnRes may exhibit additional structure.
    \item \textbf{Renormalised interventions.} Our ablations test reliance on routing pathways, not deletion of represented information. A pathway whose ablation has no measurable effect may still carry information that other pathways substitute for.
    \item \textbf{Modest absolute accuracy.} Both checkpoints score in the mid-$0.5$ range; results are best read as controlled probes.
\end{itemize}

\section{Conclusion}

Block AttnRes makes depth routing explicit. We have argued that this architectural exposure is necessary but not sufficient for interpretability, and that the right way to read it is from causal intervention on routing pathways rather than from average routing weights. On a controlled retrieval task, the same probes recover a near-degenerate single-source policy from a recency-bias-wrapped vanilla baseline whose routing pattern matches the schedule's analytic prediction, and three localised stage-specific motifs from a Block AttnRes checkpoint trained from scratch. Across the populated motifs, mass and causal importance dissociate sharply. Block AttnRes' practical value for mechanistic interpretability is not that it makes routing automatic, but that it makes routing-shaped questions answerable when --- and only when --- the routing has been part of training.

\paragraph{Future work.} (i) Whether the same stage-specific pattern reproduces on multi-hop retrieval or algorithmic tasks. (ii) Intermediate architectural conditions (AttnRes fine-tuned from vanilla; AttnRes trained with routing frozen) to disentangle architecture from training. (iii) The training trajectory: at what point does the recency-bias-like solution give way? (iv) Regularisers that encourage routes whose causal importance, not just average mass, is high. (v) Whether the \texttt{prev\_completed} pathway is genuinely unused or whether stronger probes (e.g.\ targeted activation patching) detect signal there.


\section*{Impact Statement}
This paper presents work whose goal is to advance the field of Machine Learning, specifically the mechanistic interpretability of Transformer-style models. There are many potential societal consequences of mechanistic interpretability research, none of which we feel must be specifically highlighted here.

\bibliography{example_paper}
\bibliographystyle{icml2025}


\newpage
\appendix
\onecolumn

\section{Synthetic Retrieval Dataset}
\label{app:dataset}

\paragraph{Key and value pools.} Keys are drawn without replacement from a fixed pool of 26 common English first names (\textit{Alice, Bob, Carol, \ldots, Zane}). Values are drawn without replacement from a 43-element pool spanning four loosely typed categories: colours (\textit{red, blue, \ldots}), shapes (\textit{circle, square, \ldots}), nature words (\textit{river, forest, falcon, \ldots}), and short tokens (\textit{alpha, beta, 1, 2, \ldots, 9}). Mixing categories ensures distractors cannot be solved by category-level priors alone.

\paragraph{Example construction.} For each example we sample $n \in \{2,3,4\}$ key--value pairs uniformly at random, then sample a query key uniformly from the $n$ presented keys. The correct answer is the value paired with the query key. Three distractors are drawn preferentially from other values present in the same prompt; if fewer than three in-prompt distractors are available we top up from the global value pool, excluding the correct answer.

\paragraph{Prompt styles.} To reduce surface-form sensitivity we cycle through three prompt templates in fixed proportion ($1{:}1{:}1$): \textit{plain} (\texttt{Key: value} lines and a final \texttt{Query:}), \textit{sentence} (\texttt{X is associated with Y.}), and \textit{table} (\texttt{Use the table below.}). All four candidate values are presented as continuations with a leading space, scored independently via conditional log-probability under the model.

\paragraph{Scoring.} A prediction is correct iff the model assigns the highest total log-probability to the correct continuation among the four. Random seeds are fixed for reproducibility. The full $200$-example dataset is generated in a single pass and used identically across all conditions and both checkpoints.

\section{Source Families and Routing Ablations}
\label{app:ablations}

\paragraph{Disjoint source-family definitions.} For a sublayer call with source-axis length $S$, we resolve each family symbol to a (possibly empty) set of source indices:
\begin{itemize}\setlength{\itemsep}{0.2em}
    \item \texttt{embedding}: the first source, defined for $S \ge 1$;
    \item \texttt{current}: the last source, defined for $S \ge 2$;
    \item \texttt{prev\_completed}: the source immediately before the partial block, defined for $S \ge 3$ (so it is always distinct from \texttt{embedding});
    \item \texttt{nonlocal}: completed-block sources excluding the embedding source and the most recent completed block, defined for $S \ge 4$.
\end{itemize}
At sublayer calls where the targeted family is empty for the current $S$, the ablation is a no-op at that call. The four families form a partition of the source axis when $S \ge 4$; for smaller $S$ the union is a strict subset. We additionally use \texttt{completed} (all completed-block sources) for the \texttt{keep\_only\_completed} intervention.

\paragraph{Mask-and-renormalise intervention.} Given a target family $\mathcal{F}$, we form a binary mask $m \in \{0,1\}^S$ with $m_s = 0$ iff $s \in \mathcal{F}$, and replace $\alpha$ with $\tilde{\alpha}_s = m_s\alpha_s / \sum_{s'} m_{s'}\alpha_{s'}$. \texttt{keep\_only}-style interventions invert the mask. Renormalisation preserves the total mass entering the sublayer; what changes is the model's reliance on the targeted pathway.

\paragraph{Selective application.} A single ablation can be restricted along three orthogonal axes: source family, sublayer $\in\{\textsc{attn}, \textsc{mlp}\}$, and layer subset (early/middle/late thirds). When a layer subset is specified, we floor-divide the layer count by three: early is the first $\lfloor L/3 \rfloor$ layers, middle the next $\lfloor 2L/3 \rfloor - \lfloor L/3 \rfloor$, late the remainder.

\section{Checkpoints, the Recency-Bias Schedule, and Output Differences}
\label{app:checkpoints}

We use two publicly released checkpoints from the  repository \footnote{https://github.com/wdlctc/open-attention-residuals}, both at $0.6$B parameters and identical Qwen3 backbone scale ($d{=}1024$, $L{=}28$, $h{=}16$, $h_{kv}{=}8$, $\text{ff}{=}3072$).

\paragraph{Baseline.} A vanilla Qwen3 (\texttt{Qwen3ForCausalLM}, standard additive residuals) trained from scratch with the same data and pipeline as the AttnRes condition. We load this checkpoint through \texttt{Qwen3AttnResForCausalLM} for evaluation. In every one of the $28$ layers, the per-sublayer query projection has zero norm and the recency bias is $\beta=3.0$ (codebase default). By Equation~\ref{eq:recency}, this makes the depth-softmax a deterministic function of the source-axis length and the fixed scalar.

\paragraph{AttnRes.} A Block AttnRes Qwen3 of identical scale, trained from scratch with eight blocks. Per-sublayer projections and biases are part of optimisation. The per-sublayer projection norm is non-zero in every layer for both sublayers and varies across layers, consistent with content-dependent routing.

\paragraph{Recency-bias prediction.} With $\beta=3.0$, the per-call mass on \texttt{current} is $e^\beta/(S{-}1+e^\beta)$, ranging from $0.953$ at $S=2$ to $0.742$ at $S=8$. For the $L=28$, $B=8$ model the layers partition into seven groups of four, each producing eight sublayer calls at a fixed source-axis length $S \in \{2, \ldots, 8\}$, so the seven $S$-values are equally weighted in the average. The arithmetic mean of $e^\beta/(S{-}1+e^\beta)$ over $S \in \{2, \ldots, 8\}$ is $0.840$. The observed mass on \texttt{current} of $0.840$ matches this analytic value to three decimal places.

\paragraph{Output difference between wrapped and vanilla forward passes.} The recency-bias schedule is not output-preserving: at finite $\beta$ the routing weights are a non-degenerate convex mixture, so each sublayer's input is a small but non-zero blend of earlier sources rather than the unmixed previous state expected by a vanilla PreNorm forward pass. Loading the baseline checkpoint as standard \texttt{Qwen3ForCausalLM} and as \texttt{Qwen3AttnResForCausalLM} and running both on three short prompts in matched precision, we observe a maximum absolute logit difference of $1.875$ and a mean of $\approx 0.21$. The wrapped baseline therefore behaves close to but not identically with the vanilla forward pass; what matters for our control is that its routing weights are content-independent.

\section{Routing-Mass Composition (Visual)}
\label{app:mass-figure}

\begin{figure}[h]
\centering
\includegraphics[width=0.95\linewidth]{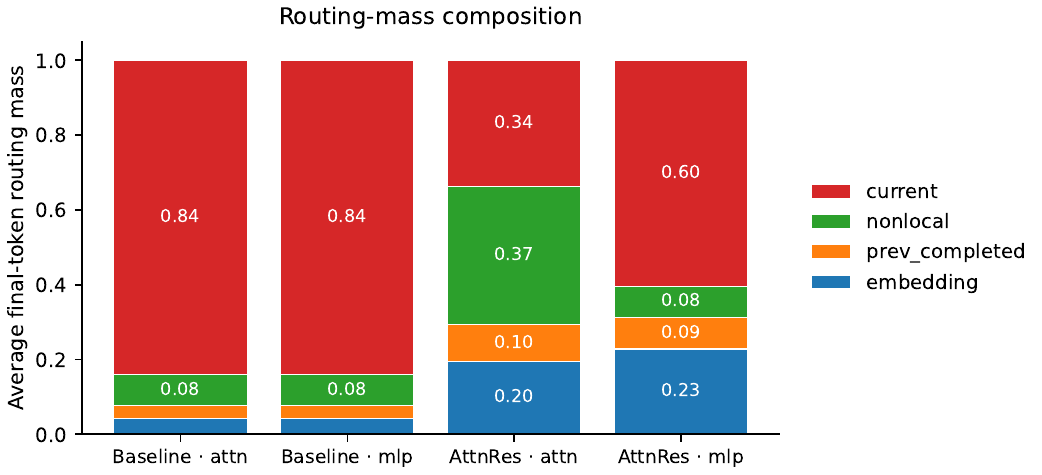}
\caption{Routing-mass composition by sublayer. The baseline allocates $\approx 0.84$ to \texttt{current} in both sublayers; the AttnRes attention sublayer is broadly distributed with the largest slice on \texttt{nonlocal}, while the AttnRes MLP sublayer is current-heavy with non-trivial \texttt{embedding} and \texttt{prev\_completed} slices.}
\label{fig:massbars}
\end{figure}

\section{Full Layer-Binned Ablation Tables}
\label{app:layerbin}

\begin{table}[h]
\centering
\footnotesize
\setlength{\tabcolsep}{4pt}
\caption{Full layer-binned sublayer ablations on the recency-bias-wrapped baseline (mean correct-answer logprob drop). All non-trivial signal is concentrated in current-attention and current-MLP routing in the early third of the stack.}
\label{tab:layerbin-base-full}
\begin{tabular}{llrrr}
\toprule
Source family & Sublayer & Early & Middle & Late \\
\midrule
embedding       & attn &  0.001 &  0.002 &  0.002 \\
embedding       & mlp  & -0.001 &  0.002 &  0.002 \\
prev\_completed & attn &  0.007 & -0.010 &  0.001 \\
prev\_completed & mlp  &  0.003 &  0.001 &  0.001 \\
current         & attn &  2.714 &  0.041 & -0.038 \\
current         & mlp  &  7.367 &  0.303 &  0.318 \\
nonlocal        & attn & -0.002 &  0.014 &  0.037 \\
nonlocal        & mlp  & -0.001 & -0.011 & -0.005 \\
\bottomrule
\end{tabular}
\end{table}

\begin{table}[h]
\centering
\footnotesize
\setlength{\tabcolsep}{4pt}
\caption{Full layer-binned sublayer ablations on the AttnRes checkpoint (mean correct-answer logprob drop). \texttt{embedding} effects are confined to early-MLP; \texttt{current} effects are early, with attention exceeding MLP; \texttt{nonlocal} effects are late-attention only; \texttt{prev\_completed} is below noise everywhere.}
\label{tab:layerbin-attnres-full}
\begin{tabular}{llrrr}
\toprule
Source family & Sublayer & Early & Middle & Late \\
\midrule
embedding       & attn & 0.140 & -0.003 & -0.000 \\
embedding       & mlp  & 6.381 & -0.000 &  0.002 \\
prev\_completed & attn & 0.045 & -0.040 &  0.008 \\
prev\_completed & mlp  & 0.011 &  0.008 &  0.050 \\
current         & attn & 2.321 & -0.120 &  0.085 \\
current         & mlp  & 1.122 &  0.494 &  0.116 \\
nonlocal        & attn & 0.076 & -0.079 &  1.729 \\
nonlocal        & mlp  & 0.005 &  0.018 & -0.202 \\
\bottomrule
\end{tabular}
\end{table}

\end{document}